\documentclass[conference]{IEEEtran}
\IEEEoverridecommandlockouts
\usepackage{cite}
\usepackage{amsmath,amssymb,amsfonts}
\usepackage{algorithmic}
\usepackage{graphicx}
\usepackage{textcomp}
\usepackage{subfigure}
\usepackage{color}
\usepackage{xcolor}
\def\BibTeX{{\rm B\kern-.05em{\sc i\kern-.025em b}\kern-.08em
    T\kern-.1667em\lower.7ex\hbox{E}\kern-.125emX}}

\begin{document}

\title{ChartNet: Visual Reasoning over Statistical Charts using MAC-Networks}
\author{\IEEEauthorblockN{Monika Sharma$^{a}$, Shikha Gupta$^{b}$, Arindam Chowdhury$^{a}$, Lovekesh Vig$^{a}$ }
\IEEEauthorblockA{\textit{$^{a}$TCS Research, New Delhi, India, Email : \{monika.sharma1, chowdhury.arindam1, lovekesh.vig\}@tcs.com} \\
\textit{$^{b}$School of Computing and EE, Indian Institute of Mandi, H.P., India, Email : shikha$\_$g@students.iitmandi.ac.in}\\
}
}

\maketitle

\begin{abstract}
Despite the improvements in perception accuracies brought about via deep learning, developing systems combining accurate visual perception with the ability to reason over the visual percepts remains extremely challenging. A particular application area of interest from an accessibility perspective is that of reasoning over statistical charts such as bar and pie charts. To this end, we formulate the problem of reasoning over statistical charts  as a classification task using MAC-Networks to give answers from a predefined vocabulary of generic answers. Additionally, we enhance the capabilities of MAC-Networks to give chart-specific answers to open-ended questions by replacing the classification layer by a regression layer to localize the textual answers present over the images. We call our network \emph{ChartNet}, and demonstrate its efficacy on predicting both in vocabulary and out of vocabulary answers. To test our methods, we generated our own dataset of statistical chart images and corresponding question answer pairs. Results show that ChartNet consistently outperform other state-of-the-art methods on reasoning over these questions and may be a viable candidate for applications containing images of statistical charts. 

\end{abstract}


\section{Introduction}
\label{sec:intro}

Statistical charts are often used to ease the understanding of large volumes of data and relationships between different elements of the data. Fully sighted humans can easily analyze the large amounts of information presented in a concise and concrete format in the charts. Many applications are built assuming this capability without taking into consideration users who may be visually challenged.
There exists a plethora of tools and algorithms capable of converting text-to-speech which can help visually impaired individuals in reading and understanding documents. However, a common stumbling block occurs in understanding and analyzing information from charts embedded in the documents. Existing machines cannot interpret and make inferences from the data visualization charts or plots because captions / tags for chart images are often very generic and do not contain any information regarding what is being represented by them. This requires systems that are able to perceive accurate visual information from statistical charts and thereafter to reason over the visual percepts. This motivates us to formulate the task of extracting information from statistical charts as a visual reasoning problem, where given a question regarding the content and relationship between elements of charts, a system should be able to answer a question by reasoning over its visual percepts. This would enable visually impaired individuals to easily understand the documents containing charts and interactively extract the statistical information embedded therein.

\begin{figure*}	
\centering  
	
	\subfigure[]{\includegraphics[width=8cm,height=5.2cm]{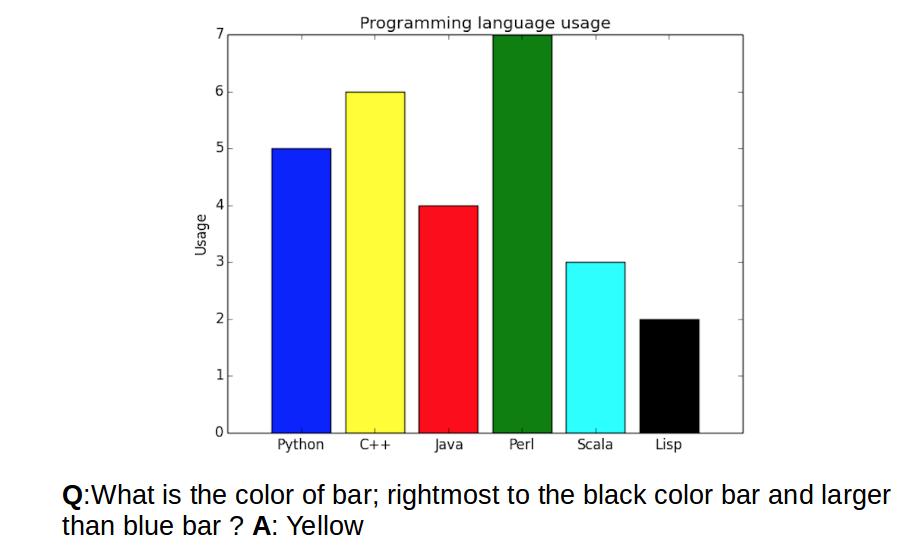}}
	\subfigure[]{\includegraphics[width=8cm,height=5.1cm]{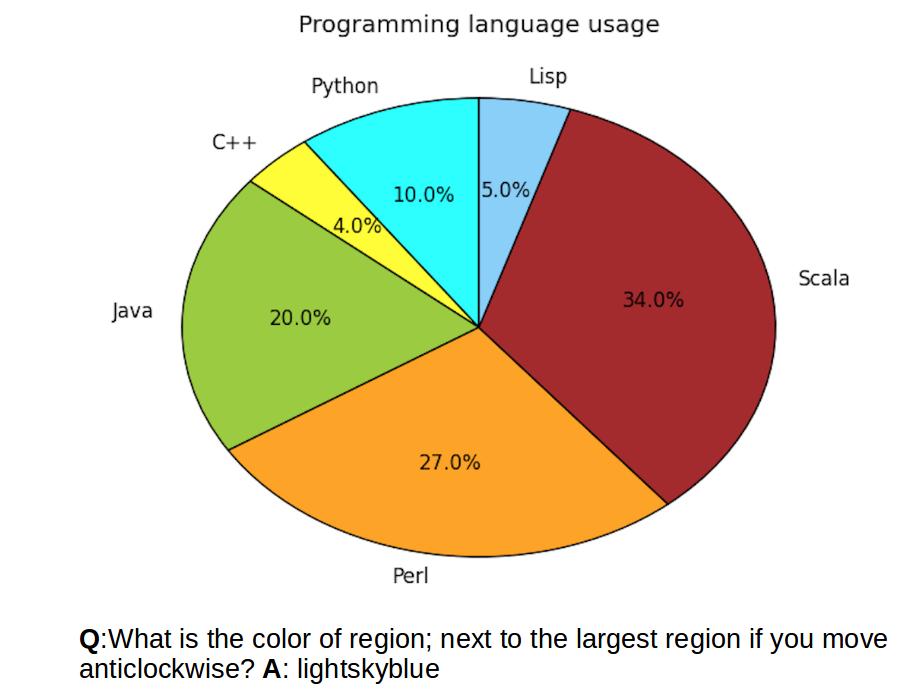}}
	\caption{Examples of bar and pie charts with corresponding question-answer pairs.}
	\label{fig:charts-examples}
\end{figure*}

We apply a variation of MAC-Networks~\cite{hudson2018compositional}, which have recently yielded promising results in visual reasoning tasks on the CLEVR dataset~\cite{johnson2017clevr}, to reason over statistical charts like bar and pie charts. The MAC-Network is an end-to-end differentiable neural network composed of Memory, Attention and Composition (MAC) cells responsible for performing a series of reasoning steps for solving a complex problem. MAC cells are recurrent networks similar to LSTM with two hidden states - control and memory. The control state contains the identity of the reasoning operation that should be performed while the memory state stores the intermediate results computed in the recurrent reasoning process so far. In this paper, we train MAC-Networks for answering questions based on charts which would enable visually impaired individuals to query information from them easily. We evaluate MAC-Networks capability for reasoning over a self-created dataset of bar and pie charts with corresponding question-answer pairs as we were unable to find suitable visual reasoning dataset for chart images. There exists FigureQA~\cite{figureqa} dataset which consists of question answers on chart images. However, FigureQA dataset contains only two types of generic answers - yes / no and does not have questions with chart specific answers. Therefore, we created our own statistical chart datasets consisting of bar and pie charts with corresponding question-answer pairs containing generic answers and bounding box annotations of chart specific textual answers present over the images. An example of a query from a bar chart is shown in Figure~\ref{fig:charts-examples}. To answer a query \emph{"What is the color of bar; rightmost to the black color bar and larger than blue color bar"} in Figure~\ref{fig:charts-examples}, the algorithm starts by finding the bar \emph{rightmost to the black color bar}, then shifts its attention to finding \emph{rightmost bar larger than blue color bar} by implicitly analyzing the heights of the bars based on y-axis and reaches to the answer "\emph{Yellow} / \emph{C++}".

Currently, MAC cells combine the information from the knowledge base generated from the image and the question embeddings to generate answer from a predefined list of generic answers. However, a fixed vocabulary of answers is not sufficient to process questions that require chart specific answers i.e., provide answers that the network has not seen during training. To mitigate this issue, we replaced the classification layer of MAC-Networks by a regression layer to give a bounding box around the text of the answer in the chart image and we named the proposed network as \emph{ChartNet}. The text present inside the generated bounding box can subsequently be read via an Optical Character Recognizer such as Tesseract~\cite{tesseract} or Google Text Vision API~\footnote{Google's Cloud Vision API : https://cloud.google.com/vision/} to give answers unique to a particular chart. We evaluate the proposed network on a self created dataset of chart images with the corresponding question-answer pairs. Finally, we demonstrate that our proposed ChartNet gives excellent performance when compared with state-of-the-art methods available for visual reasoning tasks.

\begin{figure*}[t]
\begin{center}
   \includegraphics[width=0.8\linewidth]{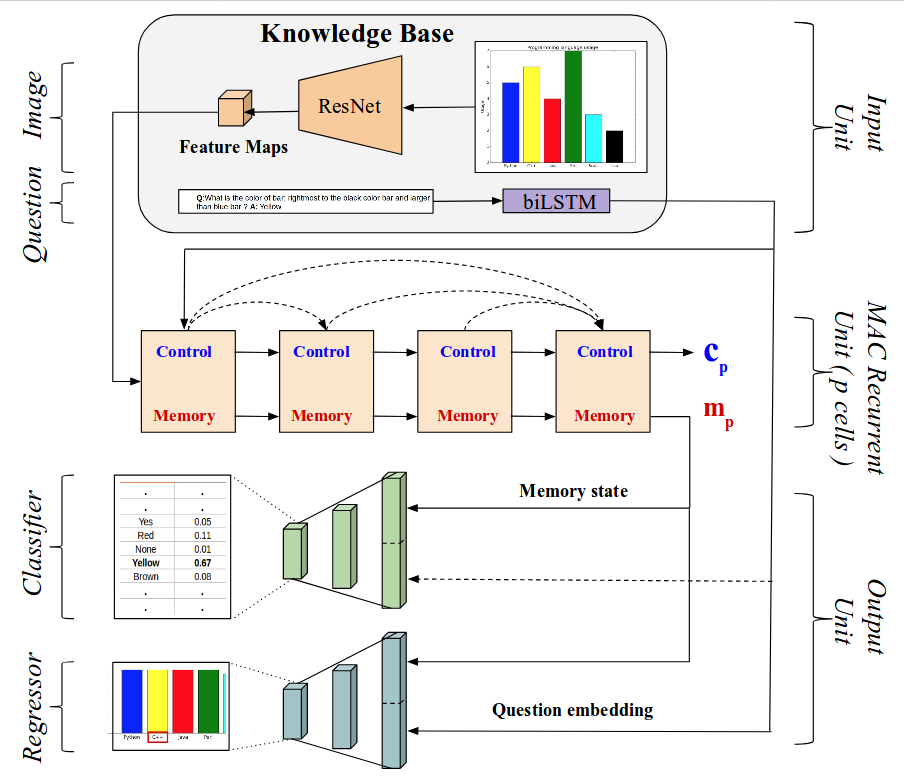}
\end{center}
   \caption{Flowchart showing proposed architecture of \emph{ChartNet} for visual reasoning over bar and pie charts. The Knowledge base consists of visual feature maps extracted using a ResNet-101~\cite{resnet101} pre-trained model. The question is encoded using a Bidirectional LSTM. A recurrent MAC layer is used to generate the reasoning output at each step, based on the question and two fully connected branches perform classification over a generic set of answers and regress the coordinates of image specific answers.}
\label{fig:flowchart-mac}
\end{figure*}

To summarize, we make following contributions in this paper :
\begin{itemize}
\item We train MAC-Networks for reasoning over statistical bar and pie charts  and to predict answers from a predefined vocabulary of generic answers.
\item We formulated the reasoning task using MAC-Networks as a regressor over textual components present in chart images, as shown in Figure~\ref{fig:flowchart-mac}, which enables MAC-Networks to give answers to open ended questions specific to a chart. We called the proposed network  \emph{ChartNet}. This alleviates the issue of having a fixed vocabulary of answers and enables the network to give answers unique to a chart image.
\item We created datasets of bar and pie charts with corresponding question and answer pairs containing generic answers and bounding box annotations of chart specific textual answers present over the images, as discussed in Section~\ref{sec:dataset}, for evaluating the effectiveness of ChartNet and comparison against state-of-the-art methods. We intend to publish these datasets for the benefit of research community.

\end{itemize}

The remaining paper is organized as follows: Section~\ref{sec:related} explores prior work done in the field of statistical chart understanding, visual question answering and visual reasoning and discusses some state-of-the-art methods. Subsequently, we give an overview of MAC-Networks and how they can reason over statistical charts in Section~\ref{sec:mac-network}. Section~\ref{sec:dataset} describes how the bar and pie chart datasets are created. The details of training of ChartNet, quantitative evaluation of ChartNet for visual reasoning over statistical charts and their comparison against state-of-the-art methods is given in Section~\ref{sec:results-and-discussions}. Finally, we conclude the paper and discuss future avenues in Section~\ref{sec:conclusion}. 

\section{Related Work}
\label{sec:related}
 
The problem of understanding statistical charts has been previously studied in the literature~\cite{charts_1,charts_2},~\cite{charts_3, charts_4} for tasks which often focus on chart detection~\cite{cdetect_1, cdetect_2}, chart classification~\cite{cClassify_1, cClassify_2}, detection and recognition of textual components~\cite{Chen2015DiagramFlyerAS, 6093320} and information extraction from charts~\cite{6093320, cClassify_2}. We found that mostly computer-vision based techniques have been used to extract visual elements from the bar charts. Chen et al.~\cite{Chen2015DiagramFlyerAS} proposed a search engine called DiagramFlyer to index a large amount of scientific charts extracted from PDFs. Their method extracted various textual components from the charts and classified their roles as x-axis, y-axis and used them along with the figure metadata. But we did not find any method in this paper to extract the original data contained in graphical components of the chart. The ReVision system proposed in ~\cite{cClassify_2} recovers the raw data encoded in bar and pie charts by extracting the labels with OCR and using a scaling factor to map from image space to data space. All these systems are highly dependent on the accuracy of the OCR. Interactive SIGHT~\cite{Sight} provides a brief initial summary that conveys the underlying message of the bar chart along with the chart's most significant features. It is currently implemented as a browser extension and works on simple charts. This program aids visually impaired individuals by providing them with the high-level knowledge that one would gain from viewing graphics in electronic documents. In this paper, we aim to automatically interpret the knowledge contained in bar and pie charts to answer questions regarding chart elements and their relationships. This task requires extracting visual information from charts and reasoning over the detected chart elements, similar to Visual Question Answering (VQA).

VQA requires co-reasoning over both image and textual questions to generate an answer. In the past, most VQA tasks have been implemented by following a pipeline which consists of learning visual features of an image using convolutional neural networks (CNN)~\cite{cnn, resnet101} and using long short term memory (LSTM)~\cite{lstm_1, lstm_2} networks based question embeddings. These two features are combined together to generate an answer as described in the baseline proposed in the VQA dataset~\cite{vqa_dataset, joint_1, joint_2}. In recent years, numerous VQA algorithms have been proposed which have achieved substantial improvements over baseline models. One example is attention mechanisms~\cite{att_1, att_2} which learn to focus on the most relevant sub-regions of the image instead of concentrating on the entire image and help to efficiently generate an answer to a question. Stacked attention methods~\cite{san_1} and co-attention~\cite{co_att} between question and image further improves the performance of VQA tasks. However, VQA tasks learn a direct mapping between inputs and outputs and do not explicitly reason to generate an answer to a question and hence, tend to learn data biases~\cite{bias_1, bias_2, johnson2017clevr}. In our case, visual reasoning over statistical charts is related to VQA with a notable difference that VQA aims to generate answers by formulating the problem as a classification task and selecting an answer to a visual question from a predefined answer list. Hence, VQA suffers from the limitation of not being able to give chart-specific textual answers present on chart images. 

To address this limitation, visual reasoning augmented models came into existence which facilitate the reasoning processes in such VQA tasks. For example, models such as neural Turing machines~\cite{neural_1, neural_2} and memory networks~\cite{memory_1, memory_2}. Module networks are also an example of reasoning augmented models which use a syntactic parse of a question to derive the architecture of a network~\cite{module_1, module_2}. After this, a set of neural modules is designed to solve a particular sub-task. However, such modular networks suffer from the propagation of error from one neural module to rest of the modular network and hence, these networks heavily depend on the correctness of the structured layout.

To mitigate these issues, very recently MAC-Networks~\cite{hudson2018compositional} were proposed. MAC-Networks are fully differentiable architectures that perform structured and explicit reasoning by sequencing a new recurrent Memory, Attention and Composition (MAC) cell. MAC moves away from monolithic black-box neural architectures towards a design that encourages
both transparency and versatility. MAC-networks solves a problem  by decomposing it into a series of attention-based reasoning steps, each performed by a MAC cell. MAC-cells have shown improved performance on visual reasoning task on CLEVR~\cite{johnson2017clevr} dataset by achieving state-of-the-art $98.9\%$ accuracy. In this paper, we propose to use standard MAC-networks architecture for the task of visual reasoning over statistical charts and formulate the problem as a classification task to give answers from a predefined vocabulary of generic answers. However, to give answers to open-ended questions about chart content and their relationships which require the capability of reading relevant text present on charts and generate answers unique to a particular chart, we augment the MAC-networks with the regression layer in place of classification layers and call our proposed network as \emph{ChartNet}. This regression feature gives MAC-networks the ability to regress over chart images to find the bounding box of the correct textual answer which can then be read by an OCR.

\section{ChartNet}
\label{sec:mac-network}
In this work, we propose to use compositional attention networks~\cite{hudson2018compositional} called MAC-Networks for bar chart and pie-chart reasoning. MAC-Networks perform iterative and structured reasoning to answer complicated visual questions (VQA). MAC networks include input unit, recurrent mac cell and output unit. We pose visual reasoning problem over statistical charts as a classification task to give answers from a predefined vocabulary of generic answers. In addition, we augment the MAC-networks with the regression layer in place of classification layers to give answers to open-ended questions about chart content and their relationships which require the capability of reading relevant text present on charts and generate answers unique to a particular chart. We named our proposed network as \emph{ChartNet}. This regression feature gives MAC-networks the ability to regress over chart images to find the bounding box of the correct textual answer to a question.

\begin{itemize}
	\item \emph{Input unit}: In the MAC-network for bar and pie chart reasoning, inputs consist of bar or pie chart images and corresponding reasoning questions. Images are first given as input to a pre-trained ResNet101~\cite{resnet101} deep CNN architecture for conv4 layer feature maps extraction, as shown in Figure~\ref{fig:flowchart-mac}. For an input image $\mathcal{I}$, the resulting obtained collection of features (tensor) are further processed by a two convolutional layer CNN architecture with depth d. The knowledge base is then defined as $\mathcal{K}^{W \times H \times D} = \{k^d_{h,w}|^{H,W}_{h,w=,1,1}\}$, where $H = W = 14$ represent the height and width of the processed image. The question string is initialized by a sequence of learned word embeddings and is further processed by the biLSTM. The question embedding is then defined as $q = [ \overleftarrow{cw_1},\overrightarrow{cw_S} ]$ where $q$ is the concatenation of the LSTM's forward and backward hidden states.  
	
	\item \emph{MAC cell}: The MAC cell is a recurrent cell designed to learn fundamental reasoning operations and implement them. It consists of three neural units namely, read, write and control that modify its dual hidden states - control and memory to perform atomic reasoning operations at each step. Therefore, at each step $i = 1 \dots p$, the $i^{th}$ the hidden state consists of the control state $c_i$  which represents the reasoning operation to be performed at the current timestep, and the memory state $m_i$ which encodes the intermediate result after reasoning. The following operations are performed on them : 
	\begin{itemize}
		\item \emph{Control} unit determines the reasoning operation to be performed at each timestep $i$ and updates the control state $c_i$. It selectively attends to the question embeddings $cw_1 \dots cw_S$ and uses the position-aware representation $q_i$ and prior control output $c_{i-1}$ to modify the current control state. 
				
		\item \emph{Read} unit retrieves relevent information $r_i$ for the $i^{th}$ reasoning step by inspecting the knowledge base $k^d_{h,w}$. The relevence is by an attention distribution $rv_i$ that assigns a probability to each item in the knowledge base which is calculated using the current control state $c_i$, representing the reasoning operation to be performed and the prior memory state $m_{i-1}$ which encodes the last intermediate result. 
		
		\item \emph{Write} unit computes the intermediate reasoning output in the $i^{th}$ timestep and combines it with memory state $m_i$. Specifically, it integrates the information retrieved from the read unit $r_i$ with previous intermediate output $m_{i-1}$, guided by the present control state $c_i$.
	\end{itemize}
	\item \emph{Output Unit}: As shown in Figure~\ref{fig:flowchart-mac}, the output unit is composed of two fully connected networks which share the common MAC backbone, for performing the dual task of regressing the coordinates of the image-specific answers and predicting the answer from a vocabulary in case of generic answers. As each intermediate memory state encodes the reasoning output at that step, a concatenation of the same with the question embedding is used to generate the answers. This concatenated vector forms the input for both the regressor and classifier networks. While the classifier network predicts a probability distribution over the pre-defined set of generic answers through a softmax normalization, the regressor network generates a $4$-dimensional vector, through a sigmoid non-linearity, in which each dimension represents a coordinate value, normalized in the range $0$ to $1$. Both networks are trained separately using a categorical cross-entropy loss for classification task and mean square error for regression task, respectively.  
\end{itemize}



\section{Dataset}
\label{sec:dataset}
We created our own synthetic datasets of bar and pie-charts for visual reasoning purposes. The bar charts dataset consists of vertical bars and was created by varying the height and number of colors of bars. The colors were chosen from a predefined list of colors. We created 10 question-answer pairs for each bar chart image. The questions are designed to test the algorithms capability of answering  questions based on the overall structure understanding of charts, retrieving data from chart images by extracting individual components from chart images and reasoning over them which requires the algorithm to collect information from multiple components and perform operations on the perceived information. The dataset follows the general Visual Question-Answering setup but we also saved bounding box annotations of textual content present over the chart images for training MAC-Networks to generate answers unique to a particular chart by learning to regress over bounding boxes of textual answers. Some examples of questions asked for chart images are as follows :
\begin{itemize}
\item What is the color of the highest bar ?
\item What is the color of bar; left to the largest bar ?
\item Does there exist a green color bar ?
\item What is the color of the bar; just right to the red color bar?
\item What is the color of the bar; rightmost to the black color bar and larger than blue bar ?
\end{itemize}

Similarly, a dataset for pie-charts was created by varying the angles and colors of sectors. Example questions for pie-charts are given below :
\begin{itemize}
\item What is the color of the largest region in pie chart ?
\item What is the color of the smallest region in pie chart ?
\item What is the color of the region; next to the largest region if you move anti-clockwise?
\item Does there exist a yellowgreen color region ?
\end{itemize}

Both the datasets consist of 2000, 500 and 500 chart images for training, validation and testing, respectively. So, total number of image question pair examples for training, validation and testing are 20000, 5000 and 5000 respectively for each dataset ( i.e., bar and pie charts). 


\section{Experimental Results and Discussions}
\label{sec:results-and-discussions}
In this section, we discuss the various experiments performed to measure the performance of ChartNet on visual reasoning over statistical charts. First, we give the details of training of ChartNet in Section~\ref{subsec:training-details}. Subsequently, we present the baselines against which we compare the performance of application of ChartNet for visual reasoning over charts in Section~\ref{subsec:baselines}. Last, we present the results of performance comparison in Section~\ref{subsec:results}.

\subsection{Experimental Setup}
\label{subsec:training-details}
Both classifier and regression networks in ChartNet use two fully connected layers with the hidden layer having $512$ units. For training, we use a batch size of $128$ in a Tesla V100 GPU machine. We used the Adam optimizer with a learning rate of $0.00001$ and trained the network for $25$ epochs at which the best validation accuracy was achieved. The optimal number of epochs required for training of MAC-Networks is determined by optimizing the validation loss.

%

\subsection{Baseline Networks}
\label{subsec:baselines}

We evaluate the performance of \emph{ChartNet} on visual reasoning tasks and compare against following three baselines.
\begin{itemize}
\item \emph{LSTM}~\cite{johnson2017inferring} : In this baseline, only questions are used to generate the answers. The questions are embedded using LSTM encoder and then the answer is predicted by a multi-layer perceptron (MLP) that has 1024 units and a softmax output which gives probability distribution over answers.
\item \emph{CNN + LSTM}~\cite{johnson2017inferring} : This method uses a convolutional neural network (CNN) to extract visual features of chart images and an LSTM to embed textual features of questions. Both the features are then combined and passed to an MLP which predicts answer to the question.
\item \emph{CNN + LSTM + SA}~\cite{johnson2017clevr} : In this baseline, the image and questions are embedded using CNNs and LSTMs respectively. The concatenated features are then passed through two rounds of soft spatial attention; a linear transform of the attention output predicts the answer.
\end{itemize}

\subsection{Results}
\label{subsec:results}
Next, we present the experimental results of the ChartNet when used as a classifier to select answers from a predefined set of generic answers for visual reasoning over charts. We use the classification accuracy as a performance measure which is defined as the fraction / percentage of the questions correctly answered. As we can see in Table~\ref{tab:accuracy-pie}, if we use only LSTM features of questions to generate the answer, the accuracy is very low (42.77\%). However, on combining LSTM embeddings of questions with visual features of chart images extracted using CNNs, the accuracy improves by 25\% which is further increased on augmenting CNN and LSTM features with a stacked attention network. The attention network learns to focus on most relevant regions  of the images for answering a question  and boosts the classification accuracy to 80.68\%. When using ChartNet for visual reasoning over chart images to give generic answers, we found that it surpasses the state-of-the-art and gives an accuracy of 91.42\%. This proves that MAC-Networks are suitable for visual reasoning over statistical charts.

  	\begin{table}[!h]
		\begin{center}
		\caption{Comparison of classification accuracy (CA) in \% for pie-chart reasoning using ChartNet as a classifier.}
		\begin{tabular}{|c|c|}
			\hline
			Model       & Accuracy \\ \hline
			LSTM \cite{johnson2017inferring}       & 42.77    \\ \hline
			CNN+LSTM  \cite{johnson2017inferring}  & 67.46    \\ \hline
			CNN+LSTM+SA \cite{johnson2017clevr} & 80.68    \\ \hline
			\textbf{ChartNet}     & \textbf{91.42}    \\ \hline
		\end{tabular}
		\label{tab:accuracy-pie}
	\end{center}
	\end{table}

Next, we discuss the performance comparison of MAC-Networks against the baselines mentioned in Section~\ref{subsec:baselines} for bar-charts. Here, as it is evident from Table~\ref{tab:accuracy-bar} that ChartNet is superior in performance as compared to baselines and give a classification accuracy of 98.14\%.

	\begin{table}[!h]
	\begin{center}
		\caption{Comparison of classification accuracy (CA) in \% for bar-chart reasoning using ChartNet as a classifier.}
		\begin{tabular}{|c|c|}
			\hline
			Model       & Accuracy \\ \hline
			LSTM~\cite{johnson2017inferring}     & 41.78   \\ \hline
			CNN + LSTM~\cite{johnson2017inferring} & 86.95   \\ \hline
			CNN + LSTM + SA~\cite{johnson2017clevr}  & 89.19    \\ \hline
			\textbf{ChartNet} & \textbf{98.14}    \\ \hline
		\end{tabular}
		\label{tab:accuracy-bar}
	\end{center}
\end{table}
	
We also evaluated the performance of ChartNet when used as a regressor to predict bounding boxes of textual answers present on images to give answers to open-ended questions. The performance measure used is Intersection over Union overlap (IOU) for bounding box regression. We obtained mean IOU for test-set as 0.84. To present results in terms of classification accuracy, we used a lower threshold of 0.8 on IOU which means that if IOU of bounding box of an answer for a question is greater than 0.8, we say that the answer is correct. Using this assumption, we obtained classification accuracy of 91.2\% on test-set for out-of-vocabulary answers.

\section{Conclusion and Future Work}
\label{sec:conclusion}
An automatic method for visual reasoning over statistical charts like bar and pie charts often found in business reports, research papers etc. is useful to human analysts / visually impaired individuals because it helps in drawing inferences from various reports by providing textual information about the figures. In this paper, we present \emph{ChartNet}, a novel architecture based on  a MAC-Network appended with a regression output to perform visual question answering over statistical charts. The model demonstrates promising performance on a visual reasoning dataset of bar charts and pie charts, and allows for prediction over unseen answers. In the future, ChartNet may be enhanced by training over multiple additional charts in different orientations and to answer questions that require numerical operations and also to generate a textual summary of the statistical charts.

\bibliographystyle{IEEEtran}
\bibliography{mac_networks}

\begin{thebibliography}{10}
\providecommand{\url}[1]{#1}
\csname url@samestyle\endcsname
\providecommand{\newblock}{\relax}
\providecommand{\bibinfo}[2]{#2}
\providecommand{\BIBentrySTDinterwordspacing}{\spaceskip=0pt\relax}
\providecommand{\BIBentryALTinterwordstretchfactor}{4}
\providecommand{\BIBentryALTinterwordspacing}{\spaceskip=\fontdimen2\font plus
\BIBentryALTinterwordstretchfactor\fontdimen3\font minus
  \fontdimen4\font\relax}
\providecommand{\BIBforeignlanguage}[2]{{%
\expandafter\ifx\csname l@#1\endcsname\relax
\typeout{** WARNING: IEEEtran.bst: No hyphenation pattern has been}%
\typeout{** loaded for the language `#1'. Using the pattern for}%
\typeout{** the default language instead.}%
\else
\language=\csname l@#1\endcsname
\fi
#2}}
\providecommand{\BIBdecl}{\relax}
\BIBdecl

\bibitem{hudson2018compositional}
D.~A. Hudson and C.~D. Manning, ``Compositional attention networks for machine
  reasoning,'' \emph{CoRR}, vol. abs/1803.03067, 2018.

\bibitem{johnson2017clevr}
J.~Johnson, B.~Hariharan, L.~van~der Maaten, L.~Fei-Fei, C.~L. Zitnick, and
  R.~B. Girshick, ``Clevr: A diagnostic dataset for compositional language and
  elementary visual reasoning,'' \emph{2017 IEEE Conference on Computer Vision
  and Pattern Recognition (CVPR)}, pp. 1988--1997, 2017.

\bibitem{figureqa}
\BIBentryALTinterwordspacing
S.~E. Kahou, A.~Atkinson, V.~Michalski, {\'{A}}.~K{\'{a}}d{\'{a}}r,
  A.~Trischler, and Y.~Bengio, ``Figureqa: An annotated figure dataset for
  visual reasoning,'' \emph{CoRR}, vol. abs/1710.07300, 2017. [Online].
  Available: \url{http://arxiv.org/abs/1710.07300}
\BIBentrySTDinterwordspacing

\bibitem{tesseract}
\BIBentryALTinterwordspacing
R.~Smith, ``An overview of the tesseract ocr engine,'' in \emph{Proceedings of
  the Ninth International Conference on Document Analysis and Recognition -
  Volume 02}, ser. ICDAR '07.\hskip 1em plus 0.5em minus 0.4em\relax
  Washington, DC, USA: IEEE Computer Society, 2007, pp. 629--633. [Online].
  Available: \url{http://dl.acm.org/citation.cfm?id=1304596.1304846}
\BIBentrySTDinterwordspacing

\bibitem{resnet101}
K.~He, X.~Zhang, S.~Ren, and J.~Sun, ``Deep residual learning for image
  recognition,'' in \emph{Proceedings of the IEEE conference on computer vision
  and pattern recognition (CVPR 2016)}, June 2016, pp. 770--778.

\bibitem{charts_1}
\BIBentryALTinterwordspacing
R.~A. Al-Zaidy and C.~L. Giles, ``Automatic extraction of data from bar
  charts,'' in \emph{Proceedings of the 8th International Conference on
  Knowledge Capture}, ser. K-CAP 2015.\hskip 1em plus 0.5em minus 0.4em\relax
  New York, NY, USA: ACM, 2015, pp. 30:1--30:4. [Online]. Available:
  \url{http://doi.acm.org/10.1145/2815833.2816956}
\BIBentrySTDinterwordspacing

\bibitem{charts_2}
\BIBentryALTinterwordspacing
S.~Elzer, S.~Carberry, and I.~Zukerman, ``The automated understanding of simple
  bar charts,'' \emph{Artificial Intelligence}, vol. 175, no.~2, pp. 526 --
  555, 2011. [Online]. Available:
  \url{http://www.sciencedirect.com/science/article/pii/S0004370210001670}
\BIBentrySTDinterwordspacing

\bibitem{charts_3}
\BIBentryALTinterwordspacing
K.~G.~S. Jagadish S~Kallimani and R.~B. Eswara, ``Extraction and interpretation
  of charts in technical documents,'' \emph{2013 International Conference on
  Advances in Computing, Communications and Informatics (ICACCI)}, 2013.
  [Online]. Available: \url{https://ieeexplore.ieee.org/document/6637202}
\BIBentrySTDinterwordspacing

\bibitem{charts_4}
\BIBentryALTinterwordspacing
J.~Poco and J.~Heer, ``Reverse-engineering visualizations: Recovering visual
  encodings from chart images,'' \emph{Comput. Graph. Forum}, vol.~36, no.~3,
  pp. 353--363, Jun. 2017. [Online]. Available:
  \url{https://doi.org/10.1111/cgf.13193}
\BIBentrySTDinterwordspacing

\bibitem{cdetect_1}
\BIBentryALTinterwordspacing
W.~Browuer, S.~Kataria, S.~Das, P.~Mitra, and C.~L. Giles, ``Segregating and
  extracting overlapping data points in two-dimensional plots,'' in
  \emph{Proceedings of the 8th ACM/IEEE-CS Joint Conference on Digital
  Libraries}, ser. JCDL '08.\hskip 1em plus 0.5em minus 0.4em\relax New York,
  NY, USA: ACM, 2008, pp. 276--279. [Online]. Available:
  \url{http://doi.acm.org/10.1145/1378889.1378936}
\BIBentrySTDinterwordspacing

\bibitem{cdetect_2}
\BIBentryALTinterwordspacing
S.~Ray~Choudhury and C.~L. Giles, ``An architecture for information extraction
  from figures in digital libraries,'' in \emph{Proceedings of the 24th
  International Conference on World Wide Web}, ser. WWW '15 Companion.\hskip
  1em plus 0.5em minus 0.4em\relax New York, NY, USA: ACM, 2015, pp. 667--672.
  [Online]. Available: \url{http://doi.acm.org/10.1145/2740908.2741712}
\BIBentrySTDinterwordspacing

\bibitem{cClassify_1}
\BIBentryALTinterwordspacing
A.~Mishchenko and N.~Vassilieva, ``Model-based chart image classification,'' in
  \emph{Proceedings of the 7th International Conference on Advances in Visual
  Computing - Volume Part II}, ser. ISVC'11.\hskip 1em plus 0.5em minus
  0.4em\relax Berlin, Heidelberg: Springer-Verlag, 2011, pp. 476--485.
  [Online]. Available: \url{http://dl.acm.org/citation.cfm?id=2045195.2045249}
\BIBentrySTDinterwordspacing

\bibitem{cClassify_2}
\BIBentryALTinterwordspacing
M.~Savva, N.~Kong, A.~Chhajta, L.~Fei-Fei, M.~Agrawala, and J.~Heer,
  ``Revision: Automated classification, analysis and redesign of chart
  images,'' in \emph{Proceedings of the 24th Annual ACM Symposium on User
  Interface Software and Technology}, ser. UIST '11.\hskip 1em plus 0.5em minus
  0.4em\relax New York, NY, USA: ACM, 2011, pp. 393--402. [Online]. Available:
  \url{http://doi.acm.org/10.1145/2047196.2047247}
\BIBentrySTDinterwordspacing

\bibitem{Chen2015DiagramFlyerAS}
Z.~Chen, M.~J. Cafarella, and E.~Adar, ``Diagramflyer: A search engine for
  data-driven diagrams,'' in \emph{WWW}, 2015.

\bibitem{6093320}
A.~Mishchenko and N.~Vassilieva, ``Chart image understanding and numerical data
  extraction,'' in \emph{2011 Sixth International Conference on Digital
  Information Management}, Sep. 2011, pp. 115--120.

\bibitem{Sight}
\BIBentryALTinterwordspacing
S.~Demir, D.~Oliver, E.~Schwartz, S.~Elzer, S.~Carberry, and K.~F. McCoy,
  ``Interactive sight into information graphics,'' in \emph{Proceedings of the
  2010 International Cross Disciplinary Conference on Web Accessibility (W4A)},
  ser. W4A '10.\hskip 1em plus 0.5em minus 0.4em\relax New York, NY, USA: ACM,
  2010, pp. 16:1--16:10. [Online]. Available:
  \url{http://doi.acm.org/10.1145/1805986.1806009}
\BIBentrySTDinterwordspacing

\bibitem{cnn}
\BIBentryALTinterwordspacing
A.~Krizhevsky, I.~Sutskever, and G.~E. Hinton, ``Imagenet classification with
  deep convolutional neural networks,'' in \emph{Proceedings of the 25th
  International Conference on Neural Information Processing Systems - Volume
  1}, ser. NIPS'12.\hskip 1em plus 0.5em minus 0.4em\relax USA: Curran
  Associates Inc., 2012, pp. 1097--1105. [Online]. Available:
  \url{http://dl.acm.org/citation.cfm?id=2999134.2999257}
\BIBentrySTDinterwordspacing

\bibitem{lstm_1}
\BIBentryALTinterwordspacing
S.~Hochreiter and J.~Schmidhuber, ``Long short-term memory,'' \emph{Neural
  Comput.}, vol.~9, no.~8, pp. 1735--1780, Nov. 1997. [Online]. Available:
  \url{http://dx.doi.org/10.1162/neco.1997.9.8.1735}
\BIBentrySTDinterwordspacing

\bibitem{lstm_2}
\BIBentryALTinterwordspacing
Y.~LeCun, B.~Boser, J.~S. Denker, D.~Henderson, R.~E. Howard, W.~Hubbard, and
  L.~D. Jackel, ``Backpropagation applied to handwritten zip code
  recognition,'' \emph{Neural Comput.}, vol.~1, no.~4, pp. 541--551, Dec. 1989.
  [Online]. Available: \url{http://dx.doi.org/10.1162/neco.1989.1.4.541}
\BIBentrySTDinterwordspacing

\bibitem{vqa_dataset}
S.~Antol, A.~Agrawal, J.~Lu, M.~Mitchell, D.~Batra, C.~Zitnick, and D.~Parikh,
  ``Vqa: Visual question answering,'' \emph{2015 IEEE International Conference
  on Computer Vision (ICCV)}, pp. 2425--2433, Dec 2015.

\bibitem{joint_1}
A.~Fukui, D.~H. Park, D.~Yang, A.~Rohrbach, T.~Darrell, and M.~Rohrbach,
  ``Multimodal compact bilinear pooling for visual question answering and
  visual grounding,'' in \emph{EMNLP}, 2016.

\bibitem{joint_2}
\BIBentryALTinterwordspacing
J.~Kim, K.~W. On, W.~Lim, J.~Kim, J.~Ha, and B.~Zhang, ``Hadamard product for
  low-rank bilinear pooling,'' \emph{CoRR}, vol. abs/1610.04325, 2016.
  [Online]. Available: \url{http://arxiv.org/abs/1610.04325}
\BIBentrySTDinterwordspacing

\bibitem{att_1}
I.~Ilievski, S.~Yan, and J.~Feng, ``A focused dynamic attention model for
  visual question answering,'' \emph{CoRR}, vol. abs/1604.01485, 2016.

\bibitem{att_2}
K.~J. Shih, S.~Singh, and D.~Hoiem, ``Where to look: Focus regions for visual
  question answering,'' \emph{2016 IEEE Conference on Computer Vision and
  Pattern Recognition (CVPR)}, pp. 4613--4621, 2016.

\bibitem{san_1}
Z.~Yang, X.~He, J.~Gao, L.~Deng, and A.~J. Smola, ``Stacked attention networks
  for image question answering,'' \emph{2016 IEEE Conference on Computer Vision
  and Pattern Recognition (CVPR)}, pp. 21--29, 2016.

\bibitem{co_att}
\BIBentryALTinterwordspacing
J.~Lu, J.~Yang, D.~Batra, and D.~Parikh, ``Hierarchical question-image
  co-attention for visual question answering,'' in \emph{Proceedings of the
  30th International Conference on Neural Information Processing Systems}, ser.
  NIPS'16.\hskip 1em plus 0.5em minus 0.4em\relax USA: Curran Associates Inc.,
  2016, pp. 289--297. [Online]. Available:
  \url{http://dl.acm.org/citation.cfm?id=3157096.3157129}
\BIBentrySTDinterwordspacing

\bibitem{bias_1}
\BIBentryALTinterwordspacing
J.~Devlin, S.~Gupta, R.~B. Girshick, M.~Mitchell, and C.~L. Zitnick,
  ``Exploring nearest neighbor approaches for image captioning,'' \emph{CoRR},
  vol. abs/1505.04467, 2015. [Online]. Available:
  \url{http://arxiv.org/abs/1505.04467}
\BIBentrySTDinterwordspacing

\bibitem{bias_2}
A.~Jabri, A.~Joulin, and L.~van~der Maaten, ``Revisiting visual question
  answering baselines,'' in \emph{ECCV}, 2016.

\bibitem{neural_1}
\BIBentryALTinterwordspacing
A.~Graves, G.~Wayne, and I.~Danihelka, ``Neural turing machines,'' \emph{CoRR},
  vol. abs/1410.5401, 2014. [Online]. Available:
  \url{http://arxiv.org/abs/1410.5401}
\BIBentrySTDinterwordspacing

\bibitem{neural_2}
A.~Graves, G.~Wayne, M.~Reynolds, T.~Harley, I.~Danihelka,
  A.~Grabska-Barwinska, S.~G. Colmenarejo, E.~Grefenstette, T.~Ramalho,
  J.~Agapiou, A.~P. Badia, K.~M. Hermann, Y.~Zwols, G.~Ostrovski, A.~Cain,
  H.~King, C.~Summerfield, P.~Blunsom, K.~Kavukcuoglu, and D.~Hassabis,
  ``Hybrid computing using a neural network with dynamic external memory,''
  \emph{Nature}, vol. 538, pp. 471--476, 2016.

\bibitem{memory_1}
\BIBentryALTinterwordspacing
J.~Weston, S.~Chopra, and A.~Bordes, ``Memory networks,'' \emph{CoRR}, vol.
  abs/1410.3916, 2014. [Online]. Available:
  \url{http://arxiv.org/abs/1410.3916}
\BIBentrySTDinterwordspacing

\bibitem{memory_2}
\BIBentryALTinterwordspacing
S.~Sukhbaatar, A.~Szlam, J.~Weston, and R.~Fergus, ``Weakly supervised memory
  networks,'' \emph{CoRR}, vol. abs/1503.08895, 2015. [Online]. Available:
  \url{http://arxiv.org/abs/1503.08895}
\BIBentrySTDinterwordspacing

\bibitem{module_1}
J.~Andreas, M.~Rohrbach, T.~Darrell, and D.~Klein, ``Learning to compose neural
  networks for question answering,'' in \emph{HLT-NAACL}, 2016.

\bibitem{module_2}
R.~Hu, J.~Andreas, M.~Rohrbach, T.~Darrell, and K.~Saenko, ``Learning to
  reason: End-to-end module networks for visual question answering,''
  \emph{2017 IEEE International Conference on Computer Vision (ICCV)}, pp.
  804--813, 2017.

\bibitem{johnson2017inferring}
J.~Johnson, B.~Hariharan, L.~van~der Maaten, J.~Hoffman, L.~Fei-Fei, C.~L.
  Zitnick, and R.~B. Girshick, ``Inferring and executing programs for visual
  reasoning,'' \emph{2017 IEEE International Conference on Computer Vision
  (ICCV)}, pp. 3008--3017, 2017.

\end{thebibliography}

\end{document}